\documentclass[10pt,conference,a4paper]{IEEEtran}
\usepackage{graphicx}
\usepackage{amsmath}

\begin{document}

\title{Learning Neural Textual Representations\\for Citation Recommendation}

\author{\IEEEauthorblockN{
Binh Thanh Kieu\IEEEauthorrefmark{1}\IEEEauthorrefmark{2},
Inigo Jauregi Unanue\IEEEauthorrefmark{1}\IEEEauthorrefmark{3},
Son Bao Pham\IEEEauthorrefmark{1}\IEEEauthorrefmark{2},
Hieu Xuan Phan\IEEEauthorrefmark{2} and
Massimo Piccardi\IEEEauthorrefmark{1}}
\IEEEauthorblockA{\IEEEauthorrefmark{1}
University of Technology Sydney, Broadway NSW 2007, Australia\\
Email: \{binh.kieuthanh, inigo.jauregiunanue\}@student.uts.edu.au,\\
\{SonBao.Pham, Massimo.Piccardi\}@uts.edu.au}
\IEEEauthorblockA{\IEEEauthorrefmark{2}
VNU University of Engineering and Technology, Vietnam National University, Hanoi\\
Email: \{binhkt,sonpb,hieupx\}@vnu.edu.vn}
\IEEEauthorblockA{\IEEEauthorrefmark{3}
Rozetta Institute, The Rocks 2000, Sydney, NSW, Australia \\
Email: inigo.jauregi@rozettainstitute.com}}

\maketitle


\begin{abstract}
With the rapid growth of the scientific literature, manually selecting appropriate citations for a paper is becoming increasingly challenging and time-consuming. While several approaches for automated citation recommendation have been proposed in the recent years, effective document representations for citation recommendation are still elusive to a large extent. For this reason, in this paper we propose a novel approach to citation recommendation which leverages a deep sequential representation of the documents (Sentence-BERT) cascaded with Siamese and triplet networks in a submodular scoring function. To the best of our knowledge, this is the first approach to combine deep representations and submodular selection for a task of citation recommendation. Experiments have been carried out using a popular benchmark dataset -- the ACL Anthology Network corpus -- and evaluated against baselines and a state-of-the-art approach using metrics such as the MRR and F1@\textit{k} score. The results show that the proposed approach has been able to outperform all the compared approaches in every measured metric.
\end{abstract}


%

\IEEEpeerreviewmaketitle

\section{Introduction}
The sustained increase in the volume of scientific publications in the past decades has made reference selection substantially more challenging, especially for inexperienced researchers or investigators who are approaching a new field. Automated citation recommendation can help ease this challenge by suggesting the most appropriate citations for a query document, e.g., a paper draft to be submitted to ICPR 2020. Most existing citation recommendation systems rank the document candidates based on their relevance to a given query, and recommend the top entries. In alternative to simple ranking, other approaches have proposed using submodular scoring functions to select the best candidates based on a trade-off between their relevance, coverage and diversity \cite{kieu-2019-submodular} or their information flow in a citation network \cite{ieee:Yu-GlobalSIP2016}. In all cases, query-based citation recommendation systems heavily rely on the effectiveness of the underlying textual representation and the scoring functions used to assess the similarity between the query and the candidates or the candidates themselves.

\begin{figure}
  \centering
  \includegraphics[width=0.95\linewidth]{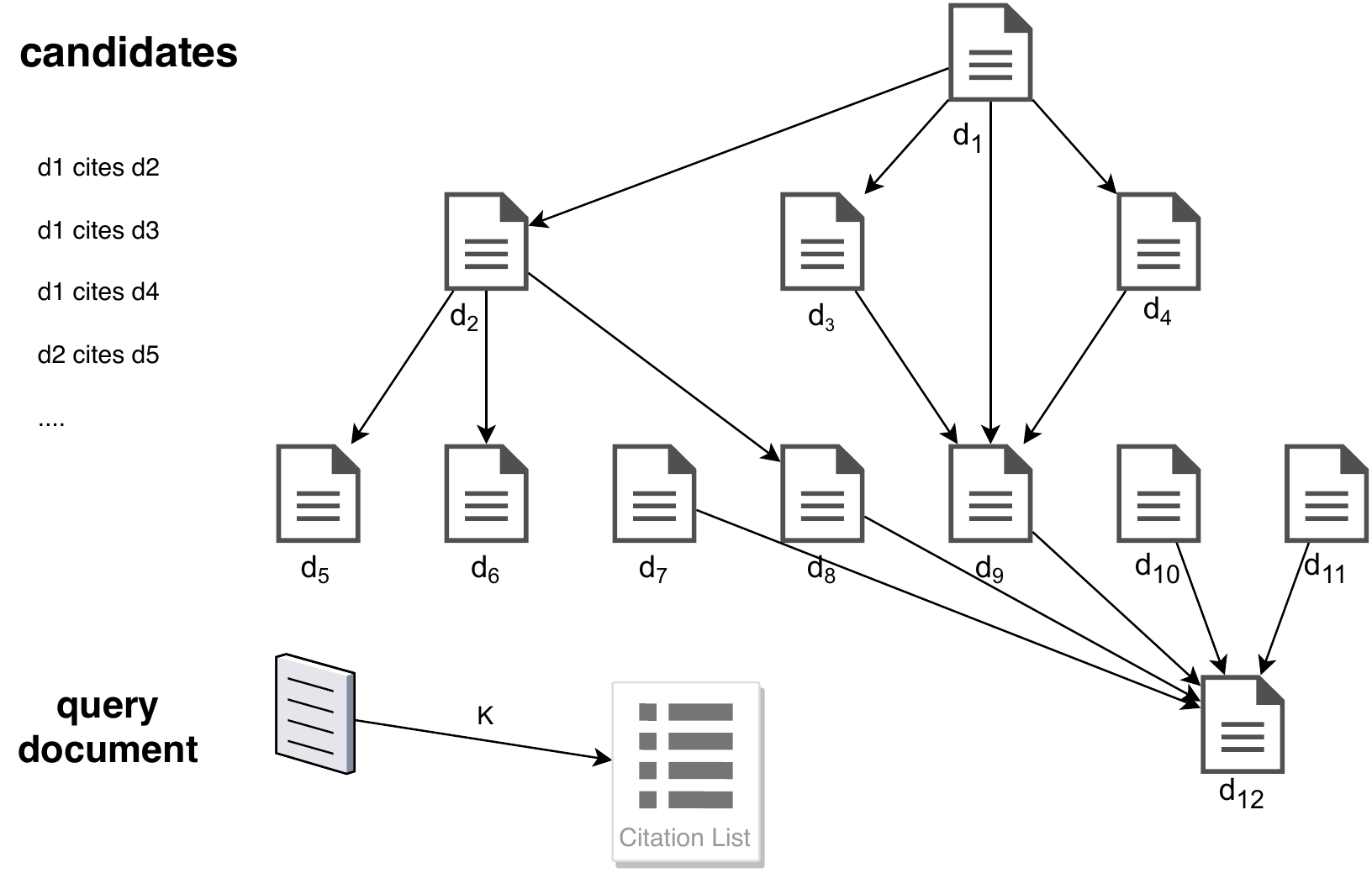}
  \caption{The citation recommendation task: a query document selects $K$ citations from a corpus of candidates organized as a citation graph.}
  \label{fig:process}
\end{figure}

Textual representation have been heavily studied as inputs to natural language understanding tasks, but they also play an important role in content-based information retrieval. Tasks in both these fields rely on textual representations that can express the semantic similarity (or, likewise, dissimilarity) between textual elements, viewed as sequences of words, word subunits or characters. Recently, pre-trained language models such as ELMo \cite{Peters-2018-elmo}, GPT-2 \cite{radford-2019-gpt2} and BERT \cite{devlin-2019-bert} have proved effective as textual representations in a broad variety of tasks. These models compute contextualized embeddings for each token which can be used as inputs for task-specific neural architectures. In this paper, we show that such models can also contribute significantly to improve the accuracy of citation recommendation.

The main contributions of our work can be summarized as:

\begin{itemize}
    
   \item[(i)] proposing representing the documents with a recently-introduced deep sequential representation (Sentence-BERT \cite{reimers-2019-sentencebert}) which allows for fast encoding and comparison (Section \ref{sec:SBERT});

   \item[(ii)] proposing fine-tuning the representation model with a training set of positive and negative examples derived from the citation graph with different strategies (Section \ref{sec:training});
   
   \item[(iii)] leveraging a submodular scoring function that balances the relevance and diversity of the citations to select the recommended list (Section \ref{sec:submod});
   
   \item[(iv)] presenting a comprehensive experimental comparison with approaches that include a strong baseline from Elasticsearch\footnote{https://www.elastic.co/} and a state-of-the-art citation recommendation approach, Citeomatic\footnote{https://github.com/allenai/citeomatic/}, on the ACL Anthology Network corpus\footnote{http://clair.eecs.umich.edu/aan/index.php}. The proposed approach has outperformed all other approaches in a range of metrics.

\end{itemize}

The rest of this paper is organized as follows: Section II presents the related work. Section III reviews feature-based document scoring and submodular selection. Section IV presents the proposed approach for deep textual representation, including the proposed fine-tuning strategies. Section V presents the experimental results, and Section VI presents the conclusion.

\section{Related Work}



From the perspective of the underlying technology, recommendation approaches can be divided into three main groups: collaborative filtering (CF) approaches, content-based filtering (CBF) approaches, and hybrid approaches. Each group has its own rationale for basing the recommendations: CF approaches focus on the recommendations or ratings of other users whose profiles are similar to the user's \cite{acm:CF-Goldberg1992}; CBF approaches compute the similarity score between keywords extracted from the user's query and text from the candidate papers \cite{wiley:Ying-AIST2014}; and hybrid approaches mix content-based and collaborative filtering techniques in various ways to improve the quality of the recommendations \cite{ieee:Bai-access2019,Chen2019,Farber2020}. 

Irrespective of the type of approach, most previous works frame citation recommendation as a ranking problem. This means that each document in the pool of candidate documents is individually scored based on the query, and the $k$ top scoring documents are chosen as the recommendations. Recently, various submodular approaches to citation recommendation have also been proposed \cite{ieee:Yu-GlobalSIP2016, kieu-2019-submodular}. Rather than simply selecting the top scoring documents, submodular approaches choose each recommendation sequentially based not only on the candidates' scores, but also on the set of documents already recommended. While more expensive computationally, submodular approaches can generate recommendation lists that are more jointly optimal; i.e., which consist of complementary, non-redundant documents in a manner more akin to the recommendations typically made by human experts. Yu et al. \cite{ieee:Yu-GlobalSIP2016} have used a submodular approach to optimize the information flow in a citation network. Kieu et al. \cite{kieu-2019-submodular} have proposed a submodular approach based on combinations of three criteria: relevance to the query, coverage of the corpus and diversity of the list. Their recommendation approach consists of two main components: 1) a document similarity function, and 2) a submodular selection. The document similarity function is used to compute a similarity score between any pair of documents (either the query and a candidate, or any two documents in the corpus). Their results have showed that Okapi BM25, a ranking function used by many search engines including Elasticsearch, has performed better than the popular TF-IDF document score. For submodular selection, they have explored a number of submodular functions, including monotone and non-monotone, and with and without meta-information. Differently from previous approaches, our approach leverages a deep textual representation of the documents, fine-tuned with a dedicated objective derived from the citation graph, and uses it for submodular selection of the recommended citations.

For what concerns textual representations, word embeddings such as word2vec \cite{Mikolov2013}, GloVe \cite{pennington-2014-glove} and fastText \cite{joulin-2016-fasttext} have been used almost ubiquitously in contemporary natural language processing (NLP). Since word embeddings are empirically compositional, they have also been used extensively to encode sentences, paragraphs and even entire documents. In the more recent years, contextualized embeddings such ELMo \cite{Peters-2018-elmo}, GPT-2 \cite{radford-2019-gpt2} and BERT \cite{devlin-2019-bert} have progressively replaced them in a number of applications. 
At a high level, all neural networks build representations of the input data as vectors/embeddings which encode useful statistical and semantic information about the input. Such latent representations are then used for performing useful downstream tasks, such as classifying an image or translating a sentence. 
In NLP, recurrent neural networks (RNNs) have historically been used to build the representations of each word in a sentence in a sequential manner. 
In turn, \textit{transformers} have gradually replaced RNNs in most NLP tasks \cite{Vaswani2017}. The transformer architecture takes a fresh approach to the representation of sequences by replacing recurrence with positional encoding. In this way, all the elements of an input sequence can be processed simultaneously to generate the output.
In addition to the effectiveness of the representations, cutting-edge libraries of  pre-trained models such as Hugging Face's Transformers \cite{wolf-2019-huggingface} are strongly contributing to their widespread adoption.

\section{Citation Recommendation Background}

In this section, we review the fundamental concepts of citation recommendation: a) the task definition, b) the scoring of the candidate documents, and c) the selection of the recommended citations.

\subsection{The Citation Recommendation Task}

Citation recommendation aims to recommend published documents as likely citations for a query document. Given a query and a corpus of published documents, $C = (d_1, d_2,...,d_N)$, the task is to choose a subset $\bar{S} \subseteq C$ to be the recommended list, $\bar{S} = \{\bar{d_i}|\bar{d_i} \in C\}$ (see Fig. \ref{fig:process}). The query can be a title, a mix of title and abstract, or an entire manuscript, with the possible addition of metadata such as authors, venues, years and other. The subset is typically chosen based on the score of a scoring function, $f(S)$, which can be defined either heuristically based on expert knowledge or using supervised learning. In the latter case, we assume the availability of a supervised training set where each document is labelled with a ground-truth citation list, $S^*$, and of a meaningful loss function, $l(S^*, S)$, which is used to evaluate the quality of a predicted list, $S$, with respect to the ground truth, $S^*$. The prediction is formally expressed as the maximization  of the scoring function, $f(S)$, under a budget constraint:
\begin{equation} 
    \begin{array}{l l}
    \bar{S} = \mathrm{argmax}_{S} & f(S) \\
    \text{s.t.} &  
    S = \{d_i|d_i \in C\} \\
                & |S| \leq K
    \end{array}
\end{equation}

\noindent where $K$ is the maximum number of citations allowed in the list. The right pane in Fig. \ref{fig:score-select} shows a depiction of this scheme. In this paper, we have set $K \in [10, 20, 50, 100]$. Please note that the selection of the optimal list may not be achieved by simply selecting each member independently, as in the general case the list needs to be jointly optimal.


\subsection{Document Scoring}

TF-IDF is likely the most common vector representation for text, where each unique term in a given text is treated as a vector dimension. Representing text as vectors allows for the straightforward use of vector operations to measure the similarity between the query and a candidate document. For instance, the scalar product or the cosine distance between the two vectors can be regarded as the relevance of the document to the query. In the TF-IDF acronym, \textit{TF} stands for term frequency, which is often just the raw count of the occurrences of each unique term in the document (many variants are possible). In turn, \textit{IDF} is the inverse document frequency, a statistic defined as the logarithm of the ratio between the number of documents in the corpus and the number of documents that contain the term, plus one. The IDF factor is used to weigh the term frequency with the ``distinctiveness'' of each unique term within the corpus.  
While still very popular in natural language processing, the TF-IDF representation has displayed some limitations when used for information retrieval. This has led to the development of alternative representations such as the BM25 (standing for Best Match 25) \cite{acm:BM25-Robertson1994}. Both TF-IDF and BM25 define the weight of each unique term in a given document as the product of some TF function and some IDF function, yet varying in the way such functions are defined.

 
In practical applications, both TF-IDF and BM25 are core components of the ranking function of Elasticsearch/Lucene, a powerful tool for document indexing and full-text search \cite{gormley2015elasticsearch}. In this paper, we adopt its definition for the TF-IDF and BM25 \textit{scores} (equations \ref{eq:es_tfidf} and \ref{eq:es_BM25}, respectively). Given a query and a candidate document, their TF-IDF score is defined as:

\begin{equation}
\label{eq:es_tfidf}
    \begin{array}{l}
        A(TF) = \sqrt{\frac{TF}{docLength}} \\ \\
        
        B(IDF) = \log{\frac{numDocs}{docFreq + 1}} \\ \\
       
        score_{TF-IDF} = \sum_{terms} A(TF) * B(IDF)
    \end{array}
\end{equation}

\noindent where \textit{TF} are the frequencies of the query terms in the candidate document, \textit{docLength} is the length of the query document, \textit{numDocs} is the total number of documents in the corpus, and \textit{docFreq} are the numbers of documents in the corpus containing each term. In turn, the BM25 score is defined as:

\begin{equation}
\label{eq:es_BM25}
    \begin{array}{l}
        A(TF) = \frac{TF * (k + 1)} {TF + k * (1 - b + b * \frac{docLength}{avgdL})} \\ \\
        
        B(IDF) = \log{\frac{numDocs - docFreq + 0.5}{docFreq + 0.5}} \\ \\
        
        score_{BM25} = \sum_{terms} A(TF) * B(IDF) 
    \end{array}
\end{equation}

\noindent where $avgdL$ is the average document length and $k$ and $b$ are two arbitrary parameters. The $avgdL$ factor is used to normalize the document's length, penalizing the score of documents longer than the average and rewarding those shorter. Parameter $k$ is used to control the term frequency's ``saturation'', i.e. to limit how much a single query term can affect the score. In turn, parameter $b$ is used to control the impact of the ratio between the document's length and the average length. If $b$ is set to zero, the ratio has no bearing on the score, while its impact increases for positive values of $b$. For $k$ and $b$ we have used Elasticsearch's default values of $1.2$ and $0.75$, respectively.


\subsection{Document Selection}
\label{sec:submod}

Most previous works on citation recommendation have treated the selection of the documents to recommend as a ranking problem, in the sense that each candidate document is scored against the query individually, and the $k$ top ranked documents are selected as the recommended list. Differently from those works, in this paper we frame the selection of the documents as the maximization of a submodular scoring function.

The concept of submodularity naturally fits the citation recommendation task and is simple to illustrate: let us call $C$ the set of all the candidate citations, $d$ an element in $C$, and $A$ and $B$ two recommendation lists (i.e., subsets of $C$). Intuitively, there will be less ``gain'' for introducing another citation into a list if such a list is already substantial. This principle is often referred to as the ``law of diminishing returns'' and can be formally expressed as:

\begin{equation}
\label{eq:subm}
A \subseteq B \rightarrow [f(B \cup d) - f(B)] \le [f(A \cup d) - f(A)]
\end{equation}

The problem of maximizing submodular functions is NP-hard and usually approximately solved via a simple, greedy algorithm which, however, enjoys theoretical guarantees for its worst-case approximation. Positing $S_0$ as the empty set, at iteration $i = 1 \ldots K$, the algorithm adds to the list the element $d \in C \setminus S_{i-1}$ that maximizes the discrete derivative $\Delta(d \mid S_{i-1}) := [f(S_{i-1} \cup d) - f(S_{i-1})]$:

\begin{equation} \label{eq:greedy_alg}
  S_i = S_{i-1} \cup \{\mathrm{argmax}_{d \in C \setminus S_{i-1}} \Delta(d \mid S_{i-1})\}  
\end{equation}








In \cite{kieu-2019-submodular}, the authors have proposed several submodular functions which balance relevance, coverage and diversity in the recommended list. Among the various functions, the one that achieved the best experimental performance is a monotone submodular function that leverages the meta-information about the authors and the venues:

\begin{equation} 
\label{eq:diversity}
    f(S) = \sum_{i = 1}^{M} \sqrt{\sum_{j \in S \cap P_i} r_{ij}}
\end{equation}

In (\ref{eq:diversity}), $P_i, i= 1 \ldots M$, are the clusters of a partition of the corpus, $C$, obtained by clustering either authors or venues, and $r_{ij} \geq 0$ is the reward for choosing recommended citation $d_j$ from the $i$-th cluster. Since the square root grows less than linearly, this function favors selecting citations from different clusters.

\section{The Proposed Approach:\\Learning Document Scoring}

Feature-based methods such as TF-IDF, BM25 and also others have proved remarkably effective for document scoring and selection. However, in recent years deep neural networks have been increasingly applied to document modeling and scoring in an end-to-end fashion. It could thus be tempting to target a scoring function for citation recommendation that can learn to optimally score the similarity of any document pair of a given training set. However, the typical corpora are large and the number of training pairs would grow too quickly. For this reason, in the following we limit our choice of similarity functions to simple, fixed functions such as the cosine and Euclidean distances, and we focus instead on learning optimal representations of individual documents. In this way, we have been able to significantly abate the size of the required training set, as explained in the following sections. 

\begin{figure*}
  \centering
  \includegraphics[width=0.9\textwidth]{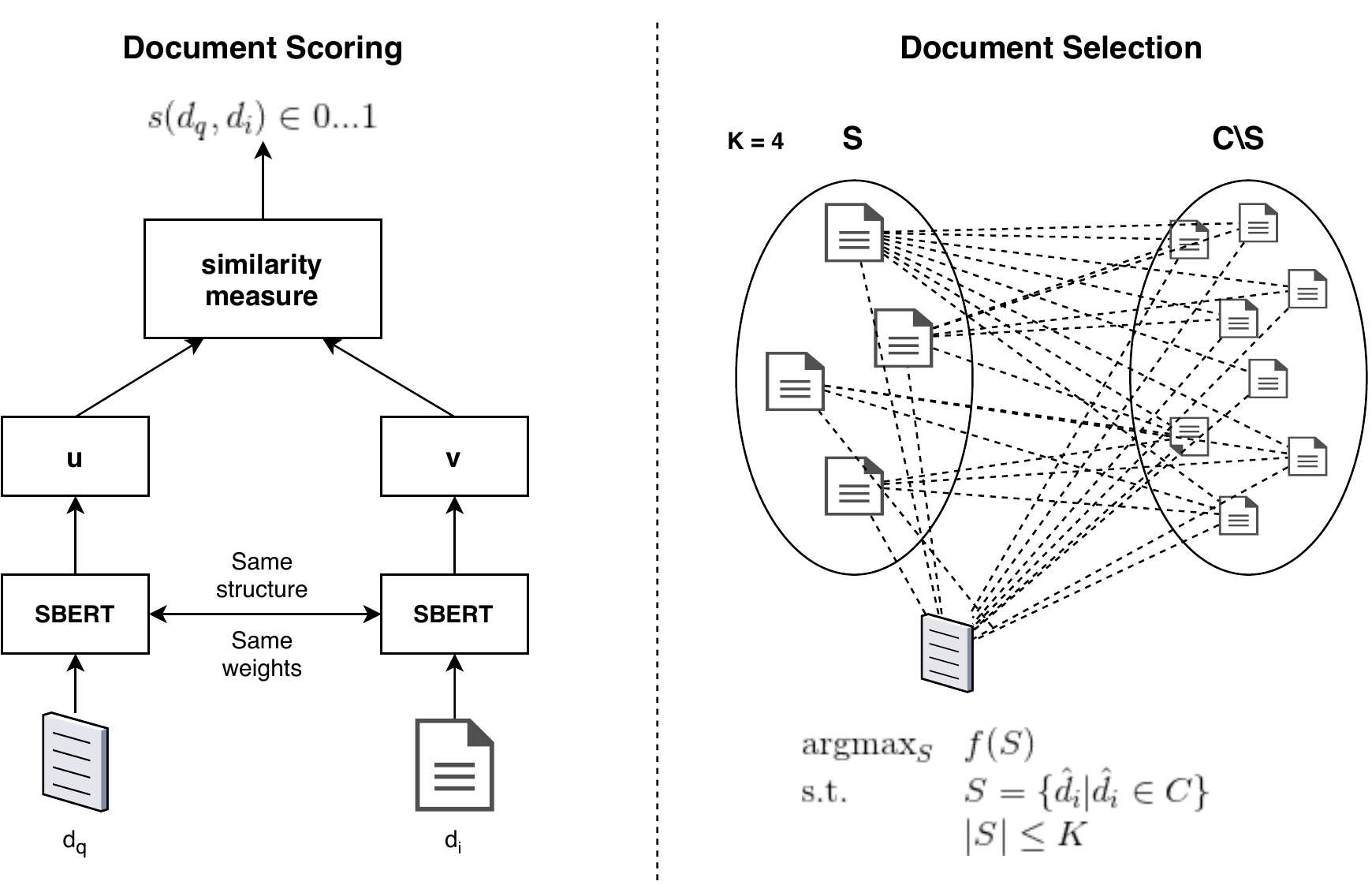}
  \caption{Left: document scoring with a Siamese network with a cosine similarity function; $d_q$: query document; $d_i$: candidate document. Right: an illustration of document selection with $K = 4$ .}
  \label{fig:score-select}
\end{figure*}

\subsection{The Deep Textual Representation}
\label{sec:SBERT} 

It is possible to use trained BERT models also to perform pairwise comparisons of documents. To this aim, BERT first uses a cross-encoder to pass the two documents to the transformer, and then uses the transformed features to produce the comparison result. However, the encoding and comparison are relatively slow -- in the order of a few milliseconds on a modern GPU -- making them impractical for any dataset of relevant size. For instance, finding the document pair with the highest similarity in a corpus of $n = 20,000$ documents (the size of the dataset used in our experiments) would  require $n*(n-1)/2 = 199,990,000$ comparisons and a total time in the order of several tens of hours.
For this reason, in this paper we have adopted \textit{Sentence-BERT} \cite{reimers-2019-sentencebert}, a model which instead embeds each document individually using any trained BERT instance as base model, and then uses fast functions such as the cosine and Euclidean distances for comparison. To further increase the speed of the model and limit its memory occupation, we have used DistilBERT, a lighter and faster version of BERT, as the base model \cite{DistilBERT}. This has abated the total time required to retrieve the most similar document pair to approximately ten seconds. 

\subsection{The Network Architecture} \label{sec:network}

Sentence-BERT \cite{reimers-2019-sentencebert}, that we refer to as SBERT for brevity in the following, uses Siamese and triplet networks to generate sentence embeddings for a variety of tasks. Both the Siamese and the triplet networks have been proven effective for scoring the similarity (or likewise, the dissimilarity) of two input sentences or paragraphs. For instance, Siamese networks with convolutional layers have been applied for matching sentences \cite{Hu-2014-Siamese-cnn}, or combined with recurrent neural networks to learn text semantic similarity \cite{Mueller-2016-Siamese-recurrent, neculoiu-2016-Siamese-recurrent}. For our work, we take advantage of the existing pre-trained models, that have been learned over very large corpora, and we fine-tune them for our citation recommendation task by providing selected samples of query and candidate documents. At run-time, the models are used to compute the score between the given query and each candidate document in the corpus.







\subsubsection{Siamese network} 
The Siamese network -- also known as the ``twin'' neural network -- is, de facto, a single network which is applied in tandem to two input vectors to produce output representations that can be compared for similarity with a simple measure such as the cosine or Euclidean distance \cite{bromley-1994-Siamese}. The transformation is provided by a non-linear feedforward network which has proved effective in a variety of tasks. To fine-tune the Siamese network, we use the Euclidean distance (or mean squared error) loss as the objective function. The left pane of Fig. \ref{fig:score-select} shows the Siamese network built on top of the SBERT encoder.

\subsubsection{Triplet network} 
The triplet network is a variant of the Siamese network which uses the same network three times to contrast the scores of positive and negative training samples \cite{hoffer-2015-triplet}. Given an anchor document $d_q$, a positive document $d^+$, a negative document $d^-$, and a similarity function between two documents $s(q,d) \ge 0$, the triplet loss tunes the network so that the similarity between $d_q$ and $d^+$ is predicted higher than that between pair $(d_q, d^-)$:

\begin{equation}
\label{eq:triplet}
\text{triplet loss} = \max[s(d_q, d^-) - s(d_q, d^+) + 1, 0]
\end{equation}

Using a simple hinge function, a loss is incurred if $s(d_q, d^+)$ is not larger than $s(d_q, d^-)$ by at least one unit. Otherwise, the loss is set to zero.




\subsection{The Training Approach}
\label{sec:training}
We refer to a positive example as a $(d_q, d^+)$ pair, where document $d_q$ is treated as the query and document $d^+$ as a known matching document, and a negative example as $(d_q, d^-)$ where document $d^-$ is known not to be matching query $d_q$. As in \cite{reimers-2019-sentencebert}, we learn the parameters of the document embedding model by using a training set of $(d_q, d^{+/-})$ pairs for the Siamese network and $(d_q, d^+, d^-)$ triplets for the triplet network.

\subsubsection{Citation graph}
To select the training samples for the supervised training, we leverage the citation graph, i.e. the graph where the nodes are the documents in the corpus and a directed edge exists between node pair $d_i$ and $d_j$ if document $d_i$ cites document $d_j$. We define a distance level, $dis(d_i, d_j)$, between nodes $d_i$ and $d_j$ as the number of nodes in the shortest path from node $i$ to node $j$ in the citation graph. For example, when document $d_i$ cites document $d_j$, $dis(d_i, d_j)$ is equal to $1$, and when $d_i$ cites a document which in turn cites $d_j$, $dis(d_i, d_j)$ is equal to $2$.

\subsubsection{Selecting the positive examples}
\label{sec:positive}
From the citation graph, we have selected the positive examples as the examples with $dis(d_i, d_j) = {1, 2, 3}$. For the MSE objective in the Siamese network, we have then defined the target similarity of the positive examples as $sim(d_i, d_j) = \theta^{dis(d_i, d_j) - 1}$ where $\theta$ is a positive constant selected using the validation set. For the triplet network objective, we have directly used the cosine distance provided by the network to enforce a margin between the distances of the selected negative and positive examples.

\subsubsection{Selecting the negative examples}
\label{sec:negative}
All the non-positive documents for a query $d_q$ have been treated as negatives, and their target 
similarity in the MSE objective has been set to $sim(d_i, d_j) = 0$. For training, we have only used subsets of the negative examples, selected with three different strategies:

\begin{itemize}

    \item \textbf{Random}: random negative documents for $d_q$.
    
    \item \textbf{Nearest}: negative documents that are closest to $d_q$ in the SBERT embedding space. The embedding space was chosen from the best model trained with random negative examples.
    
    \item \textbf{Farthest}: negative documents that are farthest from $d_q$ in the embedding space.
    
\end{itemize}

To prevent imbalance between positive and negative examples, we have set the number of negative examples for each query document to be equal to that of its positive examples.





\section{Experiments}


\subsection{Dataset}
For the experiments, we have used the ACL Anthology Network corpus (AAN), a set of $22,085$ publications in the field of computational linguistics first presented in \cite{acl:Radev-acl2013}. This dataset has been used by virtually all papers in the field of citation recommendation and is widely accepted as a benchmark. In the dataset, each document body is accompanied by a title, an abstract, a venue, authors, and a set of references that we use as ground truth for the citation recommendation task. We replicate the experimental setup of \cite{acl:bhagavatula-etal-2018} by excluding papers with no references and using the standard training ($16,128$ papers from 1960 to 2010), ``dev''/validation ($1,060$ papers from 2011) and test ($1,161$ papers from 2012) splits.

\subsection{Compared Approaches}
We have compared our approach with a strong baseline for citation recommendation, BM25, a submodular approach presented in \cite{kieu-2019-submodular}, SubRef, and a state-of-the-art method, Citeomatic \cite{acl:bhagavatula-etal-2018}. The BM25 baseline is our implementation of the popular ranking function Okapi BM25 that is used effectively and efficiently in many information retrieval systems, including Elasticsearch. SubRef uses BM25 as the similarity function for document pairs and applies a submodular search with appropriate submodular scoring functions to choose the set of recommended citations. Citeomatic is a neural approach that first embeds all documents in a learned vector space, then selects the query's nearest neighbours as candidates, and eventually reranks the candidates using a deep discriminative model to produce the final recommendation list. To train our models, we have used the hyperparameters listed in Table \ref{tab:parameters}. The learning rate used was SBERT's default, while the batch size was set to $16$ due to memory limitations. The value for $\theta$ was selected in $[0.1-0.9]$ in $0.1$ steps using the validation set. The number of epochs for all our models was set to have approximately equivalent training time. For all the compared models' parameters we have used their default values.

\begin{table}
    \begin{center}
    \begin{tabular}{l|c|c}
        \hline
        \bf Hyperparameter & \bf Siamese network & \bf Triplet network \\
        \hline
        pre-trained model & DistilBERT & DistilBERT \\
        learning rate & 2e-5 & 2e-5\\
        batch size & 16 & 16 \\
        $\theta$ & 0.4 & N/A \\
        \hline
        \bf d=1 && \\
        epochs per training & 20 & 10\\
        samples per epoch & 144,474 &  576,970 \\
       \hline
       \bf d=2 && \\
        epochs per training & 10 & 5\\
        samples per epoch & 921,422 & 7,659,898 \\ 
       \hline
       \bf d=3 && \\
        epochs per training & 5 & 3\\
        samples per epoch & 3,264,138 & 50,421,230 \\ 
        \hline
    \end{tabular}
    \end{center}
    
    \caption{\label{tab:parameters} Hyperparameters used for training the proposed approach}
\end{table}


\subsection{Evaluation Metrics}
Following \cite{acl:bhagavatula-etal-2018}, we have used the Mean Reciprocal Rank (MRR) and the F1@\textit{k} score as the evaluation metrics. The MRR records the position of the first correct citation in the predicted recommendation list, then computes its reciprocal (the higher, the better) and averages it over the test set. The F1@\textit{k} score is the harmonic mean of the corpus-level precision and recall at \textit{k}, i.e. the precision and recall in the \textit{k} top ranked citations. The precision and recall are first computed for each query document, then averaged over the test set to return the final F1 score.

\subsection{Main Results}
Table \ref{tb:general_testing_results} reports the MRR, F1@10, F1@20, F1@50 and F1@100 results for the three compared approaches and two variants of our method.
The first variant, labeled ``SBERT", only uses the Sentence-BERT model, fine-tuned as described in Section \ref{sec:training}, to embed the combined text of the title and abstract of each document. The top \textit{k} ranked documents for each query are selected as the recommendations.
The second variant, labeled ``SBERT+SubRef" uses the submodular inference of SubRef to select the recommendations based on the similarity score from SBERT. 
The results in Table \ref{tb:general_testing_results} show that even the first variant, SBERT, has been able to achieve competitive performance, with improvements over the three compared approaches in F1@50 and F1@100 and second-best results in al the other metrics. The best configuration for this model was chosen using the validation set as shown in Tables \ref{tb:training_Siamese_results} and \ref{tb:training_triplet_results}. However, the second variant, SBERT+SubRef, has reported the best results under all the metrics, with marked improvements over the runner-up (Citeomatic/Select+Rank) of $6.56$ percentage points in MRR and $3.88$, $3.67$, $3.68$ and $3.69$ points in F1@10, F1@20, F1@50 and F1@100, respectively.

\begin{table}
\begin{center}
\begin{tabular}{l|c|c|c|c|c}
\hline 
\bf Method & \bf MRR & \bf F1@10 & \bf F1@20 & \bf F1@50 & \bf F1@100 \\
\hline
\bf ElasticSearch &&&&&\\
BM25 & 0.2437 & 0.0701 & 0.0632 & 0.0446 & 0.0321 \\ 
\hline
\bf Citeomatic &&&&&\\
Select & 0.3085 & 0.1281 & 0.1339 & 0.0940 & 0.0548 \\ 
Select+Rank & 0.3777 & 0.1590 & 0.1472 & 0.0959 & 0.0549 \\ 
\hline
\bf SubRef (best) &&&&&\\
BM25-QAIv2& 0.3320 & 0.1310 & 0.1264 & 0.0911 & 0.0621\\ 
\hline
\bf SBERT (best) &&&&&\\
Siamese, d=2, farth. & 0.3493 & 0.1424 & 0.1400 & 0.1096 & 0.0797 \\ 
\hline
\bf SBERT+SubRef &&&&&\\
Siamese+QAIv2 & \bf 0.4431 & \bf 0.1978 & \bf 0.1839 & \bf 0.1327 & \bf 0.0918 \\ 
\hline
\end{tabular}
\end{center}
\caption{\label{tb:general_testing_results}Results on the test set}
\end{table}

\subsection{Discussion}





\subsubsection{Selection of the training examples}

The selection of training examples plays a critical role in the performance of the proposed approach. Clearly, it is infeasible to train with all the available negative samples as their number grows quadratically with the size of the dataset and would lead to very imbalanced training sets. Also the selection of the examples to be regarded as positives has a major impact on the performance. For this reason, in Tables \ref{tb:training_Siamese_results} and \ref{tb:training_triplet_results}
we show the results on the validation set for the Siamese and the triplet networks, respectively, for training with positive examples with distance levels = 1, 2, 3, and negative examples with the random, nearest and farthest selection strategies. The results show that the Siamese network has achieved better performance in all cases, with a best distance level of 2 (i.e., including the citations of directly-cited documents) and the farthest negative selection.

\begin{table}
\begin{center}
\begin{tabular}{l|c|c|c|c|c}
\hline 
\bf Method & \bf MRR & \bf F1@10 & \bf F1@20 & \bf F1@50 & \bf F1@100 \\
\hline
\bf Siamese, d=1 &&&&&\\
- random & 0.1986 & 0.0625 & 0.0652 & 0.0584 & 0.0482 \\  
- nearest & 0.1200 & 0.0288 & 0.0271 & 0.0222 & 0.0172 \\ 
- farthest & 0.1852 &  0.0556 &  0.0568 &  0.0520 &  0.0419  \\ 
\hline
\bf Siamese, d=2 &&&&&\\
- random & 0.4308 & 0.1662 & 0.1691 & 0.1352 & 0.0977 \\ 
- nearest & 0.1394 & 0.0319 & 0.0267 & 0.0176 & 0.0132 \\ 
- \bf farthest &\bf 0.4382 & \bf 0.1766 & \bf 0.1730 & \bf 0.1332 & \bf 0.0950  \\ 

\hline
\bf Siamese, d=3 &&&&&\\
- random & 0.3887 & 0.1412 & 0.1353 & 0.1036 & 0.0774 \\ 
- nearest & 0.1763 & 0.0541 & 0.0551 & 0.0456 & 0.0347 \\ 
- farthest & 0.3988 & 0.1477 & 0.1384 &  0.1073 &  0.0782  \\ 
\hline
\end{tabular}
\end{center}
\caption{\label{tb:training_Siamese_results}Results for the Siamese network on the validation set with different selections of the training examples}
\end{table}

\begin{table}
\begin{center}
\begin{tabular}{l|c|c|c|c|c}
\hline 
\bf Method & \bf MRR & \bf F1@10 & \bf F1@20 & \bf F1@50 & \bf F1@100 \\
\hline
\bf Triplet, d=1 &&&&&\\
- random & 0.0943 & 0.0278 & 0.0348 & 0.0335 & 0.0301 \\ 
- nearest & 0.0882 & 0.0179 & 0.0144 & 0.0094 & 0.0067 \\ 
- farthest & 0.0990 & 0.0260 & 0.0395 & 0.0356 & 0.0410 \\ 
\hline
\bf Triplet, d=2 &&&&&\\
- random & 0.1154 & 0.0339 & 0.0373 & 0.0332 & 0.0282 \\ 
- nearest & 0.0885 & 0.0215 & 0.0178 & 0.0134 & 0.010 \\ 
- farthest & 0.1103 & 0.0333 & 0.0382 & 0.0392 & 0.0355 \\ 
\hline
\bf Triplet, d=3 &&&&&\\
- random & \bf 0.3119 & 0.0888 & 0.0846 & 0.0679 & 0.0507 \\ 
- nearest & 0.1693 & 0.0380 & 0.0308 & 0.0205 & 0.0149 \\ 
- \bf farthest & 0.3092 & \bf 0.0986 & \bf 0.0941 & \bf 0.0743 & \bf 0.0563 \\ 
\hline
\end{tabular}
\end{center}
\caption{\label{tb:training_triplet_results}Results for the triplet network on the validation set with different selections of the training examples}
\end{table}

\subsubsection{Choice of the submodular inference function}
Kieu et. al \cite{kieu-2019-submodular} had proposed four submodular functions for the selection stage, including two non-monotone (QFRv1, QFRv2) which balanced relevance to the query, coverage of the corpus and non-redundancy of the list, and two monotone (QAIv1, QAv2) that balanced relevance and diversity of venues and authors. We have re-implemented these functions and evaluated their accuracy over our documents (i.e. the concatenations of the title and abstract of the original publications). Table \ref{tb:subref_testing_results} shows that monotone function QAIv2 which balances relevance and diversity of authors has achieved the best accuracy. For this reason, we have used it in conjunction with SBERT in the proposed SBERT+SubRef approach.

\begin{table}
\begin{center}
\begin{tabular}{l|c|c|c|c|c}
\hline 
\bf Method & \bf MRR & \bf F1@10 & \bf F1@20 & \bf F1@50 & \bf F1@100 \\
\hline
BM25-QFRv1 & 0.2443 & 0.0711 & 0.0639 & 0.0449 & 0.0328 \\ 
BM25-QFRv2 & 0.2428 & 0.0701 & 0.0632 & 0.0446 & 0.0321 \\ 
BM25-QAIv1 & 0.3206 & 0.1276 & 0.1218 & 0.0879 & 0.0591 \\ 
\bf BM25-QAIv2 & \bf 0.3320 & \bf 0.1310 & \bf 0.1264 & \bf 0.0911 & \bf 0.0621\\ 
\hline
\end{tabular}
\end{center}
\caption{\label{tb:subref_testing_results} Comparison of different submodular functions (BM25 as the similarity score)}
\end{table}

\section{Conclusion}


In this paper, we have proposed a novel approach to citation recommendation that leverages a deep representation of the documents. The representation has been obtained by encoding each document with Sentence-BERT (SBERT), a recently proposed transformer-based approach for text embedding. In the paper, we have proposed fine-tuning SBERT with positive and negative examples derived with various strategies from the citation graph. In addition, we have proposed performing the prediction of the recommended list with a submodular scoring function that balances the relevance of the recommended citations with the diversity of their authors. The experimental results over a benchmark dataset (the ACL Anthology Network corpus) have shown that the proposed approach has been able to outperform all the compared approaches, including a state-of-the-art neural approach, by a remarkable margin in all metrics ($3.67-6.56$ percentage points over the runner-up). In the near future, we aim to explore the integration of submodular scoring in the training stage and extend the evaluation to other domains and document types.





%
\bibliographystyle{IEEEtran}
\bibliography{icpr2020-ref-reformat}

\end{document}